\documentclass[sigconf]{acmart}

\usepackage{graphicx}
\usepackage{hyperref}
\usepackage{natbib}
\usepackage{subcaption}
\usepackage{xcolor}

\AtBeginDocument{%
  }

\copyrightyear{2026}
\acmYear{2026}
\setcopyright{cc}
\setcctype{by}
\acmConference[GeoIndustry '26]{The 5th ACM SIGSPATIAL International Workshop on Spatial Big Data and AI for Industrial Applications}{November 03--06, 2026}{Riverside, CA, USA}
\acmBooktitle{The 5th ACM SIGSPATIAL International Workshop on Spatial Big Data and AI for Industrial Applications (GeoIndustry '26), November 03--06, 2026, Riverside, CA, USA}
\acmDOI{10.1145/3849727.3856786}
\acmISBN{979-8-4007-3053-5/2026/11}

\begin{document}

\title[GeoNatureAgent (GNA)]{GeoNatureAgent (GNA): A Framework and Benchmark for Pre-Production Evaluation of Tool-Using Agents on Geospatial and Environmental Tasks}
\titlenote{This workshop paper substantially extends our SIGSPATIAL~2026 poster on the GeoNatureAgent benchmark~\cite{diazireland2026geonatureagent}; Section~\ref{sec:intro} details the new material.}


\author{Gabriel Diaz-Ireland}
\authornote{Corresponding author.}
\email{gdiazir1@jhu.edu}
\orcid{0009-0003-9049-4947}
\affiliation{%
  \institution{Universidad Cat\'{o}lica de \'{A}vila (UCAV)}
  \city{\'{A}vila}
  \country{Spain}
}
\affiliation{%
  \institution{Johns Hopkins University}
  \city{Baltimore, MD}
  \country{USA}
}

\author{Diego Prieto-Herr\'{a}ez}
\email{diego.prieto@ucavila.es}
\orcid{0000-0003-4636-2261}
\affiliation{%
  \institution{Universidad Cat\'{o}lica de \'{A}vila (UCAV)}
  \city{\'{A}vila}
  \country{Spain}}

\author{Mario Garc\'{\i}a Peces}
\email{mario.garcia.peces@gmail.com}
\affiliation{%
  \institution{Independent Researcher}
  \city{Madrid}
  \country{Spain}
}

\author{Javier Vel\'{a}zquez}
\email{javier.velazquez@ucavila.es}
\orcid{0000-0002-9188-3827}
\affiliation{%
  \institution{Universidad Cat\'{o}lica de \'{A}vila (UCAV)}
  \city{\'{A}vila}
  \country{Spain}}

\author{Benjamin Zaitchik}
\email{zaitchik@jhu.edu}
\orcid{0000-0002-0698-0658}
\affiliation{%
  \institution{Department of Earth and Planetary Sciences, Johns Hopkins University}
  \city{Baltimore, MD}
  \country{USA}
}

\author{Devika Jain}
\email{kakkar@fas.harvard.edu}
\orcid{0000-0001-7800-929X}
\affiliation{%
  \institution{Center for Geographic Analysis, Harvard University}
  \city{Cambridge, MA}
  \country{USA}
}

\renewcommand{\shortauthors}{Diaz-Ireland et al.}

\begin{abstract}
	Before tool-using LLM agents are deployed in environmental and geospatial workflows, teams need evidence that an agent reliably selects the right operations against real APIs---yet no reusable, self-hostable substrate exists for such pre-production evaluation.
	We introduce \textbf{GeoNatureAgent (GNA)}, a framework for pre-production evaluation of tool-using agents. GNA publishes a fixed sixteen-tool geospatial interface (twelve principal operations and four auxiliary) as a Model Context Protocol (MCP) server, making the agent under test the only variable: any MCP-compatible client is a drop-in subject scored against an identical tool layer, task suite, and deterministic scorer. Its flagship instance is a 103-task benchmark---a 93-task main suite across 18 categories plus a ten-task comparison expansion---spanning tool selection, multi-turn conversation, spatial reasoning, cross-indicator synthesis, error handling, multilingual understanding, and task rejection, among others, evaluated against an open, self-hostable geospatial API serving three environmental indicators across Spain and Portugal.
	We evaluate nine benchmark-tested large language models across Vertex~AI, OpenRouter, and Anthropic platforms under three temperature-1.0 seeds per model, reporting capability and per-case cost as orthogonal axes.
	Results manifest that: (1)~Claude~Sonnet~4 achieves the highest capability at $61.7\%\pm0.7\%$ on the combined 103 tasks ($60.8\%\pm0.8\%$ on the main suite), followed closely by DeepSeek~V3.2 at $57.9\%\pm3.9\%$, while no other model exceeds $53\%$; (2)~the cost-accuracy Pareto frontier is occupied mostly by open-weight models, with DeepSeek~V3.2 offering $93\%$ of Claude's capability at 11.3$\times$ lower list-price cost ($\$0.011$/case); and (3)~under strict all-checks scoring, which requires end-to-end task completion, the best model sits $24$--$36$ percentage points below the $85$--$97\%$ reported on general-purpose GIS benchmarks, whereas the top four models' per-check partial credit ($86$--$90\%$) is comparable, so much of that gap reflects scoring strictness rather than task difficulty alone.
	We further demonstrate geographic and domain extensibility by integrating BigEarthNet~V2 land-cover data for Portugal alongside Spanish CO$_2$ suitability and gully erosion indicators.
	GeoNatureAgent (GNA)---the MCP tool server, evaluation harness, benchmark, and self-hostable API---is publicly available\footnote{\url{https://github.com/gabrielireland/GeoNatureAgent_Benchmark}} and designed for reuse: swapping the tool executors and task suite instantiates an equivalent benchmark for any geospatial domain.
\end{abstract}




\maketitle

\section{Introduction}
\label{sec:intro}

Environmental monitoring at scale requires multi-temporal geospatial analysis---detecting land cover change, assessing erosion risk, computing vegetation indices, and generating statistical summaries for regulatory compliance. Practitioners routinely spend most of their effort on data wrangling, discovering datasets, reprojecting coordinate systems, and debugging code against remote sensing APIs~\cite{zhang2023geogpt, li2023llmgeo}---an expertise barrier that limits adoption by domain scientists who understand environmental systems but lack GIS programming skills.

Recent advances in large language models (LLMs) have produced agent systems that translate natural language into executable geospatial operations, from LLMs with tool pools~\citep{zhang2023geogpt, akinboyewa2024giscopilot} to fine-tuned specialists~\citep{zhang2024gtchain, zhang2025envgpt} to multi-agent systems achieving $85$--$97\%$ accuracy~\citep{luo2025geojsonagents, lee2025geollmsquad}, with heavy industry investment behind them~\citep{google2025earthai, planet2025anthropic, szwarcman2024prithvi}.

However, a critical evaluation gap persists: the few public geospatial agent benchmarks target general GIS tasks (GeoBenchX~\citep{krechetova2025geobenchx}, ThinkGeo~\citep{shabbir2025thinkgeo}) rather than environmental science workflows; knowledge benchmarks (EnviroExam~\citep{huang2024enviroexam}) evaluate knowledge, not agent behavior; code-generation benchmarks (UnivEARTH~\citep{kao2025univearth}) do not reflect the structured API interaction production systems employ ---$58\%$ of LLM-generated Earth Engine code fails to execute; and GEO-Bench/GEO-Bench-2~\citep{lacoste2023geobench, simumba2025geobench2} evaluate pixel-level vision models, not agent-level tool orchestration.

We fill this gap with \textbf{GeoNatureAgent (GNA)}, a framework that reframes agent evaluation as connecting an agent to a fixed, standards-based tool substrate and scoring its behavior deterministically. GNA contributes:
\begin{enumerate}
    \item \textbf{A reusable agent-evaluation substrate} ---a fixed sixteen-tool geospatial interface published as a Model Context Protocol (MCP) server, paired with a 103-task benchmark (93-task main suite across 18 categories + ten-task expansion) and a deterministic eight-check scorer (no LLM-as-judge). With tools, tasks, and scorer fixed, any MCP-capable agent is a drop-in subject-under-test---a reusable substrate, not a one-off harness.
    \item \textbf{Multi-seed cross-platform evaluation} ---a systematic comparison of nine LLMs spanning closed-source models and open-weight families, each evaluated under three tempe\-rature-1.0 seeds on identical infrastructure with mean-$\pm$-standard-deviation reporting; capability and per-case cost are reported as orthogonal axes rather than collapsed into a single gated score.
    \item \textbf{Cost-efficiency Pareto analysis} ---a quantitative mapping showing the Pareto frontier is mostly open-weight, with DeepSeek~V3.2 offering $93\%$ of Claude's capability at 11.3$\times$ lower list-price cost.
    \item \textbf{Reproducible MLOps pipeline and open API} ---a fully automated Cloud Build pipeline producing conversation-trace-level results, plus an Apache-2.0 self-hostable FastAPI service so the entire evaluation can be reproduced without any private endpoint.
\end{enumerate}

Throughout the results, a \emph{Build your own} note closes each subsection, naming the committed code path that reproduces the analysis and scales it to new models, tasks, indicators, and regions.

\paragraph{Relation to prior publication.} This paper substantially extends our SIGSPATIAL~2026 poster~\cite{diazireland2026geonatureagent}, which introduced the benchmark. New beyond that four-page prior publication are: the framework-first formulation and the MCP tool substrate (Section~\ref{sec:framework}), the nine-model leaderboard (adding GPT-4o and Gemma-3-27B), the v6 close-value expansion (Section~\ref{sec:v6}), the universal tool-call parser, and the expanded failure taxonomy and cross-benchmark positioning; all empirical results herein are new or re-derived.

\section{Related Work and Positioning}
\label{sec:related}

\subsection{Geospatial AI Agent Architectures}

The literature reveals four paradigms of geospatial AI agent architectures. The \textbf{LLM + Tool Pool} approach pairs general-purpose LLMs with callable tools (GeoGPT~\citep{zhang2023geogpt}, LLM-Geo~\citep{li2023llmgeo}, GIS~Copilot~\citep{akinboyewa2024giscopilot}). \textbf{Fine-tuned specialists} sacrifice generality for expertise: GTChain~\citep{zhang2024gtchain} achieved $32.5\%$ higher accuracy than GPT-4; EnvGPT~\citep{zhang2025envgpt} fine-tuned an 8B model on 100M environmental tokens, rivaling GPT-4o-mini. \textbf{Multi-agent systems} decompose tasks: GeoJSON~Agents~\citep{luo2025geojsonagents} achieved $97\%$ with a GPT-4o Planner-Worker architecture vs. $49\%$ single-agent; GeoLLM-Squad~\citep{lee2025geollmsquad} gained $17\%$ from specialized sub-agents. \textbf{Cost-efficient hybrids}: Geo-OLM~\citep{stamoulis2025geoolm} enables sub-7B models to perform within $10\%$ of GPT-4o at 100$\times$ lower cost.

\subsection{Benchmarks and Evaluation}

Agent tool-use is evaluated by GeoBenchX~\citep{krechetova2025geobenchx}, ThinkGeo~\citep{shabbir2025thinkgeo} ($486$ remote sensing tasks), and UnivEARTH~\citep{kao2025univearth} (only $33\%$ accuracy on Earth Engine code generation and a $58\%$ code failure rate). GeoBenchX is the closest comparable work: $202$ tasks across five categories, evaluated via $24$ general-purpose GIS tools with an LLM-as-judge protocol; crucially, its tools are low-level primitives operating on local files (\texttt{load\_data}, \texttt{filter\_categorical}, \texttt{make\_\allowbreak choropleth\_\allowbreak map}), whereas GNA tools are high-level domain operations against a production cloud API (Table~\ref{tab:benchmark_comparison}).

Map reasoning: MapEval~\citep{dihan2025mapeval} (no model exceeds $67\%$), MapQA~\citep{arnold2025mapqa} (humans $91\%$, models $<50\%$). Spatial reasoning: GPSBench~\citep{truong2026gpsbench}, SpatiaLab~\citep{wasi2026spatialab} (humans $88\%$, best model $55\%$). Foundation models: GEO-Bench~\citep{lacoste2023geobench} and GEO-Bench-2~\citep{simumba2025geobench2} evaluate \emph{pixel-level} classification/segmentation (\emph{can this model classify a satellite tile?}), complementary to GNA's agent-level tool orchestration (\emph{can this agent reason about which tools to call?}); PANGAEA~\citep{marsocci2024pangaea} finds that foundation models do not consistently outperform supervised baselines.

\subsection{Environmental Science AI}

Specialized models such as EnvGPT~\citep{zhang2025envgpt} demonstrate that focused fine-tuning on environmental tokens allows smaller architectures to rival general-purpose models like GPT-4o-mini. In industry, Google’s Earth~AI~\citep{google2025earthai} reaches $82\%$ accuracy on geospatial Q\&A (vs $50\%$ baseline Gemini), and MCP-grounded interfaces reach a factual precision of $4.78/5.0$ in air quality monitoring~\citep{pan2025mcpairquality}. Regulatory mandates such as Spain’s RD~214/2025~\citep{spain2025rd214}, which requires roughly $4,000$ organizations to report carbon footprints, further drive demand for automated geospatial analysis.

\subsection{Positioning}
\label{sec:positioning}

No existing benchmark evaluates environmental analysis agents performing structured tool calling against a real geospatial API with multi-turn conversation, multilingual inputs, and cross-indicator synthesis. GNA fills this gap; this paper extends our prior benchmark~\cite{diazireland2026geonatureagent} with the framework formulation, the MCP tool substrate, and the expanded empirical study enumerated in Section~\ref{sec:intro}.

Table~\ref{tab:benchmark_comparison} contrasts GNA with GeoBenchX, the closest comparable benchmark, as both share the structured tool-calling paradigm. GNA's best-model accuracy ($60.8\%$, Section~\ref{sec:results}) is notably lower than the $85$--$97\%$ reported by GeoBenchX and GeoJSON~Agents. Four factors may contribute: (1)~real API interaction vs.\ local file operations; (2)~environmental domain complexity (geographic knowledge, data limitations, display-only layers); (3)~multi-turn conversation history; and (4)~strict binary evaluation. Only the last can be tested with our own data, and it explains much of the gap: under partial credit (fraction of checks passed) the top four models reach $86$--$90\%$, comparable to the range reported elsewhere, although the comparison still controls for neither task sampling nor rubric. The strict gap therefore measures end-to-end task completion, which deployment requires, rather than per-step competence; we read the cross-benchmark comparison as indicative, and a same-protocol re-run of prior suites is future work (Section~\ref{sec:limitations}). Our accuracies align with the harder map- and spatial-reasoning benchmarks above. Against code generation, structured tool calling constrains the output space: our $60.8\%$ on real API tool calls compares favorably with UnivEARTH's $33\%$ and $58\%$ code-execution failure rate~\citep{kao2025univearth}, again across protocols but consistent with structured APIs being the safer choice over free-form code generation for production environmental systems.

\begin{table}[ht]
    \caption{Comparison of structured tool-calling geospatial benchmarks. Accuracies are not same-protocol: GeoBenchX uses an LLM judge, GNA strict all-checks scoring (best per-check partial credit: $89.6\%$).}
    \label{tab:benchmark_comparison}
    \centering
    \begin{tabular}{lcc}
        \toprule
        & \textbf{GeoBenchX} & \textbf{\shortstack[c]{GeoNatureAgent \\ Benchmark}} \\
        \midrule
        Tasks              & 202            & 103 \\
        Categories         & 5              & 18 \\
        Tools              & 24 primitives  & 12 principal ops \\
        Tool abstraction   & Low-level      & High-level \\
        Data source        & Local files    & Cloud API (COG + JSON) \\
        Domain             & General GIS    & Environmental \\
        Countries          & ---            & Spain + Portugal \\
        Indicators         & ---            & 3 \\
        Multi-turn         & No             & Yes \\
        Best accuracy      & $85$--$90\%$       & $60.8\%$ \\
        Evaluation         & LLM-as-judge   & Metric-based \\
        \bottomrule
    \end{tabular}
\end{table}

\section{The GNA Framework}
\label{sec:framework}

\paragraph{The benchmark as an MCP tool server.} To make GNA a reusable \emph{substrate} rather than a one-off harness, we expose its sixteen-tool interface as a Model Context Protocol (MCP) server~\citep{mcp}, an open standard for connecting language-model agents to external tools. The server is a thin wrapper ($\sim$95 lines) over the \emph{same} tool definitions and dispatcher used by the evaluation harness and the production API---no tool logic is reimplemented---so any MCP-capable client (Claude Desktop, an OpenAI-style agent, or our own evaluation loop) drives an identical tool layer, byte-for-byte the one the API serves. This cleanly separates the \emph{benchmark} (the fixed tools, data, and scorer) from the \emph{agent under test} (the only variable): an evaluation is simply to connect an agent to the MCP server, issue the task prompts, and score the resulting tool calls deterministically.

\paragraph{A deliberately tools-only surface.} MCP defines several server capabilities---\emph{Tools} (discover and invoke callable operations), \emph{Resources} (server-supplied readable data/context), and \emph{Prompts} (server-supplied templates/scaffolding). GNA implements \textbf{only the Tools capability} (\texttt{list\_tools} and \texttt{call\_tool}). This is central to what the benchmark measures: the competency under evaluation is \emph{tool selection}---whether an agent chooses the right tool, with the right arguments, at the right time, from natural-language intent. Exposing \emph{Resources} would let an agent read pre-baked context and bypass the tool-choice decision; exposing \emph{Prompts} would hand it scaffolding that leaks the solution path---either would contaminate precisely the capability we score. The in-process harness and the MCP transport present an identical decision surface, so results are comparable across both.

\begin{figure}[ht]
    \centering
    \includegraphics[trim={0.7cm 0.7cm 0.7cm 1.4cm}, clip, width=0.92\columnwidth]{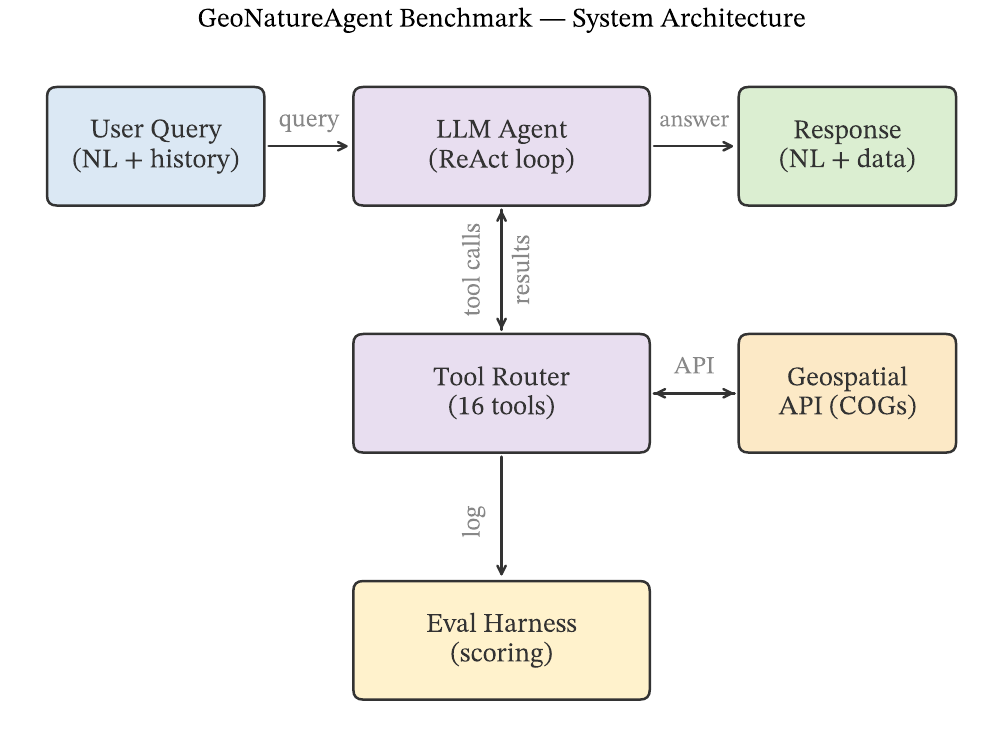}
    \caption{GNA system architecture. The LLM agent interacts with a production API serving COG data; the eval harness logs all tool calls and scores responses. NL denotes natural language.}
    \Description{}
    \label{fig:architecture}
\end{figure}

Figure~\ref{fig:architecture} shows the system architecture. The agent operates in a ReAct-style loop~\citep{yao2023react}: it receives a query, reasons about tools, executes calls, observes results, and generates a response. The agent has access to sixteen tools: the twelve principal operations in Table~\ref{tab:tools}, plus four auxiliary tools for layer discovery and erosion time-series statistics.

\begin{table}[ht]
    \caption{Principal agent tools. The first eight are domain-specific; the last four are adapted from GeoBenchX~\citep{krechetova2025geobenchx}. The agent additionally exposes four auxiliary tools not listed here (layer discovery and erosion statistics).}
    \label{tab:tools}
    \centering
    \small
    \begin{tabular}{lp{4.8cm}}
        \toprule
        \textbf{Tool} & \textbf{Purpose} \\
        \midrule
        \texttt{lookup\_province} & Resolve province $\to$ boundary geometry \\
        \texttt{lookup\_municipality} & Resolve municipality (+ province hint) \\
        \texttt{analyze\_area} & Zonal statistics for indicator in AOI \\
        \texttt{analyze\_multi\_layer} & Multi-indicator analysis in single call \\
        \texttt{compare\_areas} & Compare two areas on the same indicator \\
        \texttt{find\_top\_n} & Rank provinces by indicator value \\
        \texttt{toggle\_layer} & Enable/disable map layer visibility \\
        \texttt{generate\_chart} & Bar/stacked-bar chart from data \\
        \midrule
        \texttt{create\_buffer} & Buffer geometry by distance (km) \\
        \texttt{select\_spatial\_rel} & Select features by spatial predicate \\
        \texttt{get\_centroids} & Extract centroid coordinates \\
        \texttt{reject\_task} & Decline unsolvable tasks \\
        \bottomrule
    \end{tabular}
\end{table}

\paragraph{Multi-backend inference client.}
Fair cross-model comparison requires normalizing heterogeneous API surfaces: a native Anthropic backend and a LiteLLM backend for OpenAI-compatible endpoints normalize responses ---including tool-call representations ---into the one schema the eval harness consumes. Notably, Llama~4 models emit tool calls as Python-style function invocations rather than JSON; a dedicated AST-based parser extracts these so Llama~4~Scout is scored on the same footing as the JSON-native models.

\emph{Build your own:} the substrate is domain-agnostic---replacing the tool executors behind the MCP server (\path{api/agent/mcp_server.py}) with your own domain's operations and the task JSON with your own cases yields a benchmark for that domain, with dispatcher, ReAct loop, scorer, and aggregation unchanged.

\section{The Benchmark Instance}
\label{sec:benchmark}

GeoNatureAgent Benchmark comprises 103 tasks: a 93-task main suite organized into 18 categories (see Table~\ref{tab:categories}), spanning easy ($19\%$), medium ($47\%$), and hard ($33\%$) difficulty levels, plus the ten-task v6 expansion (Section~\ref{sec:v6}), reported both separately and in a combined 103-task metric.

\begin{table}[ht]
    \caption{GeoNatureAgent Benchmark main-suite (v5) task categories ($93$ of the $103$ tasks, 18 categories).}
    \label{tab:categories}
    \centering
    \small
    \begin{tabular}{lcp{3.9cm}}
        \toprule
        \textbf{Category} & \textbf{Tasks} & \textbf{Tests} \\
        \midrule
        Tool selection         & 21 & Correct tool choice\\
        Cross-indicator        &  8 & CO$_2$ $\times$ erosion $\times$ land cover synthesis \\
        Deep dive              &  6 & Multi-tool environmental profiling \\
        Interpretation         &  7 & Reasoning over analysis results \\
        Error handling         &  6 & Non-existent entities, invalid indicators \\
        Habitat analysis       &  7 & BigEarthNet V2 land cover (Portugal) \\
        Language               &  6 & Galician, Basque inputs \\
        Municipality           &  4 & Municipality-level analysis, disambiguation \\
        Memory (multi-turn)    &  6 & Follow-up questions, context retention \\
        Spatial reasoning      &  4 & Coastal, meseta, community knowledge \\
        Multi-muni ranking     &  3 & Multi-municipality comparisons \\
        Threshold              &  3 & Numeric threshold queries \\
        Temporal change        &  1 & Cross-country temporal context \\
        Error recovery         &  3 & Graceful fallback when data is unavailable \\
        Ranking                &  2 & Top-N and bottom-N province queries \\
        Comparison             &  2 & Close-value and directional comparisons \\
        Single analysis        &  2 & Basic single-province analysis \\
        Province aggregation   &  2 & Aggregate statistics across provinces \\
        \bottomrule
    \end{tabular}
\end{table}

Each task specifies: a natural language query (optionally with multi-turn history), expected tool calls, must-contain/must-not-contain strings, maximum rounds, cost budget, and domain-expert ground truth. Error handling tasks are deliberately unsolvable (non-existent municipalities, statistics requested from display-only layers), testing the agent's ability to decline gracefully rather than hallucinate.

\paragraph{Worked example.} Consider case \texttt{V5\_21} (comparison): \emph{``Compare CO$_2$ geological storage suitability between Murcia and C\'{o}rdoba. Which province is better suited?''} The gold solution issues a single \texttt{compare\_areas} call, which returns Murcia $68.5\%$ vs.\ C\'{o}rdoba $68.4\%$---within tolerance ($\Delta 0.1$~pp against $2.0$~pp), i.e.\ effectively tied. Every model fails this case: most runs never issue the comparison call (a \emph{tool-missing} error), and the few that do still miss on answer phrasing, not on the $0.1$-point margin---a failure of \emph{tool selection}, not numeric discrimination.

The benchmark covers three environmental indicators across two countries:

\begin{itemize}
    \item \textbf{CO$_2$ absorption suitability} (categorical, Spain): legislative pre-screening encoding five MITECO spatial criteria~\citep{spain2025rd214}, producing three classes (Not Eligible, Eligible with Conditions, Eligible). Served as a Cloud Optimized GeoTIFF (COG).
    \item \textbf{Gully erosion probability} (continuous, Europe): machine learning prediction from the LUCAS~2022 soil survey~\citep{borrelli2025gully} ($0$--$100\%$). Served as a COG.
    \item \textbf{BigEarthNet~V2 land cover} (categorical, Portugal): 7-class land cover distribution derived from 75k+ labeled Sentinel-2 patches~\citep{clasen2024reben}, aggregated by Portuguese district. Served as pre-computed JSON statistics.
\end{itemize}

Seven additional layers are available for visual display but not for statistical analysis, creating natural hallucination traps.

To align our tool surface with general-purpose GIS benchmarks, we integrate four canonical GIS operations from GeoBenchX \citep{krechetova2025geobenchx} into our toolset: proximity buffering (\texttt{create\_buffer}), feature selection by spatial relationship (\texttt{select\_\allowbreak features\_\allowbreak by\_\allowbreak spatial\_\allowbreak relationship}), centroid extraction (\texttt{get\_centroids}), and explicit task rejection (\texttt{reject\_task}) to prevent hallucination on out-of-scope queries. These are adapted from GeoBenchX's local-file paradigm to operate against the cloud API, are available across the full 103-case suite, and are invoked wherever a query calls for a spatial operation rather than through dedicated cases.

\paragraph{Design rationale.} The category structure is a capability taxonomy: each category isolates one agentic competency required for environmental geospatial analysis. Case counts are weighted by capability importance rather than balanced---tool selection receives the most cases ($21$) because it is the pivotal agentic skill (Section~\ref{sec:category}), whereas niche capabilities receive one or two. The uneven sizes are deliberate; we treat the aggregate leaderboard as the primary measure and category-level results as indicative (Section~\ref{sec:results}). Difficulty labels are author-assigned by expected tool-chain depth and are approximate.

\paragraph{Tool coverage across data domains.} Tool coverage varies by layer: the CO$_2$ and gully-erosion indicators (raster COGs, Spain/Europe) are served by the full toolset, whereas the BigEarthNet land-cover indicator (pre-computed district statistics, Portugal) is served only by \texttt{analyze\_area}, so comparison, ranking, and legend queries over the Portuguese layer require composing multiple \texttt{analyze\_area} calls rather than invoking a single convenience tool. We treat this as part of the evaluation---a capable agent must recognize when a nominally-applicable tool does not cover a layer and adapt---a realistic tool-selection challenge.

\section{Evaluation Protocol}
\label{sec:scoring}

Each task is scored by eight check types (see Table~\ref{tab:checks}). A task \emph{passes} only when \textbf{all} applicable capability checks pass (strict binary; the cost budget is reported, not gated). We additionally compute partial-credit metrics for finer-grained analysis.

\begin{table}[ht]
    \caption{Scoring checks. Binary capability pass requires every applicable check except \texttt{max\_cost\_usd} to pass.}
    \label{tab:checks}
    \centering
    \small
    \begin{tabular}{lp{5.3cm}}
        \toprule
        \textbf{Check} & \textbf{Pass condition} \\
        \midrule
        \texttt{expected\_tools} & Every expected tool was called (recall\,=\,1.0). Extra tools do not cause failure.\textsuperscript{a} \\
        \texttt{expected\_actions} & Every expected UI action was generated (recall\,=\,1.0). \\
        \texttt{must\_contain} & Each required keyword appears in the answer (case-insensitive substring). \\
        \texttt{must\_not\_contain} & Forbidden keyword does not appear in the answer. \\
        \texttt{numeric\_accuracy} & For each ground-truth entry, the label is found in the answer and the nearest percentage value is within tolerance.\textsuperscript{b} \\
        \texttt{chart\_generated} & At least one chart URL was produced (only checked when \texttt{generate\_chart} is an expected tool). \\
        \texttt{max\_rounds} & Agent loop iterations $\leq$ budget. \\
        \texttt{max\_cost\_usd}\textsuperscript{c} & Estimated token cost $\leq$ per-case budget. \emph{Reported but not gated} on capability pass/fail. \\
        \bottomrule
    \end{tabular}
    \medskip

    {\footnotesize
    \parbox{\linewidth}{
    \textsuperscript{a}Tool F1 is computed as a continuous metric but is not used as a gate.\\
    \textsuperscript{b}A 120-character window after the label is searched for the first \texttt{float\%} pattern; tolerance defaults to $\pm2$\,pp.\\
    \textsuperscript{c}Per-case budgets default to $\$0.10$ in \texttt{v5}; cost is logged for the cost-accuracy analysis but excluded from the binary capability score.}}
\end{table}

\paragraph{Continuous metrics.} Beyond binary pass/fail, we report \emph{check score} (fraction of checks passed per case), \emph{keyword coverage}, \emph{tool F1}, and \emph{cost utilisation}. Tool and action checks gate on \emph{recall} rather than F1: an agent that calls all expected tools plus an exploratory extra should not be penalised.

\paragraph{Error classification.} When a task fails, the first failing check (in priority order: \texttt{tool\_missing} $\to$ \texttt{chart\_missing} $\to$ \texttt{wrong\_data} $\to$ \texttt{rounds\_exceeded} $\to$ \texttt{cost\_exceeded} $\to$ \texttt{keyword\_missing} $\to$ \texttt{forbidden\_keyword}) determines the error category; runs that fail only on \texttt{cost\_exceeded} are excused by the capability metric, so the distribution (Table~\ref{tab:failuremodes}) reports the \emph{dominant} failure mode per run, not an exhaustive multi-label tally.

\paragraph{Capability and cost as orthogonal axes.} Per-case cost budgets are deployment thresholds, not measures of reasoning ability, so we decouple them: the binary leaderboard scores capability only, and per-case cost is a Pareto axis (Section~\ref{sec:cost}), following HELM~\citep{liang2023helm} and GAIA~\citep{mialon2023gaia}. Cost-gated numbers remain recoverable from the per-case CSV (\texttt{passed\_with\_cost\_gate}).

\section{Experiments and Results}
\label{sec:results}

\subsection{Experimental Setup}

We evaluate nine LLMs on the full 103-task GeoNatureAgent Benchmark (93-task v5 main suite, 18 categories, plus the ten-task v6 expansion; Section~\ref{sec:v6}) across Vertex~AI Model-as-a-Service, OpenRouter, and the Anthropic Messages API (see Table~\ref{tab:models}). All experiments use the v3 system prompt (a zero-shot ReAct-style instruction, committed at \path{api/agent/prompts/v3.md}), \texttt{max\_\allowbreak turns=10}, \texttt{max\_tokens=4096}, and \texttt{temperature=1.0}---chosen to expose run-to-run variance under realistic sampling noise rather than the deterministic temperature-0 regime that overstates production stability. The prompt is held constant across all models to isolate capability from prompt-engineering effort, and experiments run sequentially inside a single Cloud Run Job to ensure identical infrastructure conditions.

\begin{table}[ht]
    \caption{Models evaluated. Six via Vertex AI (five MaaS plus Gemini native), two via OpenRouter, Claude via the Anthropic Messages API.}
    \label{tab:models}
    \centering
    \begin{tabular}{llrl}
        \toprule
        \textbf{Model} & \textbf{Provider} & \textbf{Params} & \textbf{Access} \\
        \midrule
        DeepSeek V3.2 & DeepSeek AI & 671B MoE & Vertex MaaS \\
        GLM-5 & Zhipu AI & --- & Vertex MaaS \\
        Gemini 2.5 Pro & Google & --- & Vertex native \\
        Claude Sonnet 4 & Anthropic & --- & Anthropic API \\
        GPT-4o & OpenAI & --- & OpenRouter \\
        Qwen3-235B & Alibaba & 235B MoE & Vertex MaaS \\
        GPT-OSS-120B & OpenAI & 120B & Vertex MaaS \\
        Llama 4 Scout & Meta & 109B MoE & Vertex MaaS \\
        Gemma-3-27B & Google & 27B & OpenRouter \\
        \bottomrule
    \end{tabular}
    \medskip

    {\footnotesize\parbox{\linewidth}{Vertex MaaS models run on provider-managed serving stacks whose quantization and region are neither disclosed nor controlled by us; absolute cross-provider latency and cost carry this confound. The tool interface is identical across all models, so capability differences reflect the model rather than the tool layer.}}
\end{table}

\paragraph{Aggregation.} To control for run-to-run variance under temperature-1.0 sampling, every model runs the full case set under three seeds $\{42, 1337, 2024\}$ where the API exposes a \texttt{seed} parameter; the Anthropic API does not, so Claude~Sonnet~4 is evaluated under two independent temperature-1.0 samples. Reported accuracies are the per-seed mean $\pm$ one standard deviation. Every result row carries its seed and case id, and a deterministic script compiles the committed CSVs that are the single source of truth for every number in this paper.

\subsection{Overall Results}

\begin{table}[ht]
    \caption{Capability leaderboard (accuracy in \%, combined rank): nine models on the combined 103-task suite and the 93-task v5 main suite (Section~\ref{sec:v6}). Per-seed mean $\pm$ one sd, decoupled from cost (Section~\ref{sec:scoring}); cost is the mean per-seed main-suite run total, \$/case the per-call mean.}
    \label{tab:leaderboard}
    \centering
    \small
    \begin{tabular}{rlrrrr}
        \toprule
        \textbf{\#} & \textbf{Model} & \textbf{All (103)} & \textbf{Main (93)} & \textbf{Cost} & \textbf{\$/case} \\
        \midrule
        1 & Claude Sonnet 4\textsuperscript{$\dagger$} & $61.7 \pm 0.7$ & $60.8 \pm 0.8$ & $\$11.84$ & $0.127$ \\
        2 & DeepSeek V3.2    & $57.9 \pm 3.9$ & $56.3 \pm 3.1$ & $\$1.05$  & $0.011$ \\
        3 & GLM-5            & $52.4 \pm 2.6$ & $50.2 \pm 2.2$ & $\$3.58$  & $0.038$ \\
        4 & Gemini 2.5 Pro   & $49.2 \pm 3.9$ & $48.0 \pm 3.3$ & $\$4.87$  & $0.052$ \\
        5 & Qwen3-235B       & $44.3 \pm 5.0$ & $41.2 \pm 4.3$ & $\$0.89$  & $0.010$ \\
        6 & GPT-4o           & $43.7 \pm 2.6$ & $41.6 \pm 2.7$ & $\$6.49$  & $0.070$ \\
        7 & GPT-OSS-120B     & $32.7 \pm 0.6$ & $34.1 \pm 1.2$ & $\$8.27$  & $0.089$ \\
        8 & Llama 4 Scout    & $25.9 \pm 3.1$ & $26.9 \pm 2.1$ & $\$0.32$  & $0.003$ \\
        9 & Gemma-3-27B      & $14.6 \pm 1.0$ & $15.8 \pm 1.6$ & $\$5.81$  & $0.062$ \\
        \bottomrule
    \end{tabular}
    \medskip

    {\footnotesize
    \parbox{\linewidth}{
    \textsuperscript{$\dagger$}Claude Sonnet~4 is reported from two temperature-1.0 samples (the Anthropic Messages API does not implement deterministic seeding); the other eight models are evaluated under three seeds. Claude's variance estimate therefore rests on fewer observations; a third sample was not drawn because its two samples already show the tightest spread in the group ($\pm 0.8$~pp) at the highest per-run cost (\$11.84).}}
\end{table}

Table~\ref{tab:leaderboard} summarizes the results aggregated across seeds. \textbf{Claude~Sonnet~4} leads on capability at $61.7\%\pm0.7\%$ combined ($60.8\%\pm0.8\%$ main suite), with \textbf{DeepSeek~V3.2} a close second at $57.9\%\pm3.9\%$ ($56.3\%\pm3.1\%$); GLM-5 and Gemini~2.5~Pro form the second tier, Qwen3-235B and GPT-4o are effectively tied in the third, and GPT-OSS-120B and Llama~4~Scout follow. \textbf{Gemma-3-27B} floors the leaderboard at $15.8\%\pm1.6\%$, partly a harness artifact: it emits tool calls as \texttt{```tool\_code} markdown blocks rather than the parsed structured format, and extending the parser to accept that dialect roughly tripled its accuracy from a raw $5.7\%$ and dropped its no-tool-call rate from $96\%$ to $68\%$---a weak but now fairly-measured model (Claude~Sonnet~4.5 exhibits a similar tool-call format mismatch that we flag but do not correct; Section~\ref{sec:prompt}). Only two models exceed $55\%$ strict accuracy and none reach $65\%$: on this suite, end-to-end completion of environmental tool chains remains a challenge even for the strongest models. Inter-seed standard deviations are tight ($0.8$--$4.3$ percentage points on the main suite), so tier-level ordering is robust to seed noise, though adjacent ranks are not (below).

Three findings stand out. First, \textbf{the top of the leaderboard is bimodal but close}: Claude~Sonnet~4 and DeepSeek~V3.2 are separated by only $4.5$ percentage points (Welch's $t=2.40$, $df\!\approx\!2$, $p\!\approx\!0.13$) ---within statistical noise at our sample size ---while both lead the third-place GLM-5 by more than $6$~pp ($t\!\geq\!2.76$); the top-versus-bottom gap is robust (Claude vs.\ Llama~4~Scout, $t=25.0$). Second, \textbf{accuracy tracks scale strongly within our small open-weight sample and across all three seeds}: Llama~4~Scout (109B MoE) trails Qwen3-235B, which trails DeepSeek~V3.2 (671B MoE). Third, \textbf{closed-source frontier models do not categorically dominate open-weight peers}: Claude leads only marginally above DeepSeek, and Gemini~2.5~Pro sits below both open-weight contenders---a pattern reinforced on the cost axis (Section~\ref{sec:cost}).

\paragraph{Quantifying the scale trend.} Restricting to the five open-weight models with disclosed parameter counts (27B--671B), accuracy rises monotonically with total parameters: Spearman $\rho=1.0$, Pearson $r=0.98$ on $\log$ parameters, ranking identical on each seed individually; at $n=5$ the exact two-sided permutation $p=2/5!=0.017$ clears the conventional $0.05$ threshold. Since the models differ in architecture and training, we read this as a strong within-sample trend rather than a general scaling law.

\emph{Build your own:} adding a model to this leaderboard is one experiment YAML in \path{benchmark/experiments/} pointing at any OpenAI-compatible or Anthropic endpoint; the committed aggregation scripts re-emit the leaderboard and per-category CSVs, and \path{scripts/scaling_stats.py} recomputes the scale trend as the open-weight sample grows.

\paragraph{Partial credit tells a different story.} Per-check partial credit is far higher than strict accuracy: $71$--$90\%$ across the nine models and $86$--$90\%$ for the top four. Failing cases often show partial competence---correct tools but a missing keyword---so many failures are ``near-misses'', and this scoring difference is the main measurable reason our strict accuracies sit below those of looser-scored benchmarks (Section~\ref{sec:positioning}).

\paragraph{Bootstrap confidence intervals.} To quantify how much of the inter-model spread is small-sample noise, we compute $95\%$ bootstrap confidence intervals on the accuracy of each of the seven original v5 models by resampling the $93$ main-suite cases ($2{,}000$ resamples). The intervals are wide and overlap substantially between adjacent models---Claude~Sonnet~4 $[51.1, 69.9]$ and DeepSeek~V3.2 $[47.7, 64.5]$ overlap almost entirely, as do GLM-5 $[41.2, 58.1]$ and Gemini~2.5~Pro $[39.8, 56.3]$---so adjacent-model and category-level rank claims should be read as indicative rather than definitive at $n=93$. Only the top-versus-bottom gap sits well outside any overlap: Llama~4~Scout's interval $[19.7, 33.7]$ lies entirely below Claude~Sonnet~4's $[51.1, 69.9]$.

\subsection{Cost-Accuracy Analysis}
\label{sec:cost}

\begin{figure}[ht]
    \centering
    \includegraphics[trim={0.1cm 0.1cm 1.8cm 1.9cm}, clip, width=0.94\columnwidth]{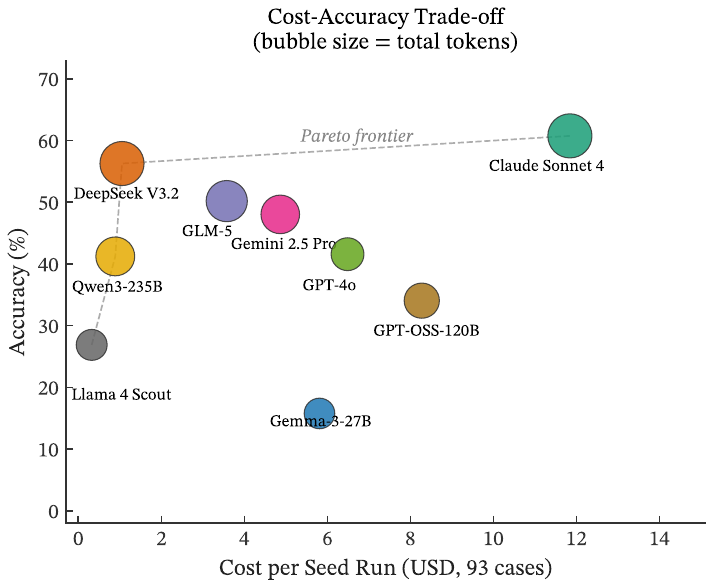}
    \caption{Cost-accuracy trade-off (bubble size $\propto$ total tokens). The Pareto frontier (dashed) runs Scout~$\to$~Qwen3-235B~$\to$~DeepSeek~V3.2~$\to$~Claude~Sonnet~4; three of the four are open-weight.}
    \Description{}
    \label{fig:cost_accuracy}
\end{figure}

Cost per case varies by more than an order of magnitude, from $\$0.003$ for Llama~4~Scout to $\$0.127$ for Claude~Sonnet~4 (Table~\ref{tab:leaderboard}). The cost-accuracy Pareto frontier (Figure~\ref{fig:cost_accuracy}) comprises four models: \textbf{Llama~4~Scout} (cheapest, $26.9\%$ capability), \textbf{Qwen3-235B} ($41.2\%$ at $\$0.010$/case), \textbf{DeepSeek~V3.2} ($56.3\%$ at $\$0.011$/case), and \textbf{Claude~Sonnet~4} (highest capability at $60.8\%$, $\$0.127$/case). Three of the four are open-weight; the remaining five models---closed-source Gemini~2.5~Pro and GPT-4o, open-weight GLM-5, GPT-OSS-120B, and Gemma-3-27B---are dominated. The DeepSeek~V3.2 entry is particularly notable: it delivers $93\%$ of Claude's capability ($56.3\%$ vs.\ $60.8\%$) at $11.3\times$ lower per-case cost, making it the strongest single-model recommendation for cost-sensitive deployments. Costs are list prices on provider-managed serving stacks whose hardware and quantization we cannot observe (Table~\ref{tab:models}), so cross-provider ratios reflect what a deployer pays, not intrinsic model efficiency.

Projected to a production deployment serving $1,000$ queries/day (a straight extrapolation of the per-case cost in Table~\ref{tab:leaderboard}), the capability leader (Claude~Sonnet~4, $60.8\%$) costs $\$127$/day, while the second-place DeepSeek~V3.2 ($56.3\%$) costs $\$11.3$/day ---a $4.5$-percentage-point capability gap for an $11.3\times$ price difference. For most deployments the right operating point sits on the open-weight Pareto frontier, unless the absolute capability ceiling matters more than throughput economics; the pattern echoes Geo-OLM's cost-efficiency findings~\citep{stamoulis2025geoolm} among general-purpose models.

\paragraph{From cost to carbon.} Extending the cost analysis to environmental footprint, we estimate energy and carbon per query from token usage (blended $0.40$~Wh per $1{,}000$ tokens; global-average grid intensity $400$~gCO$_2$/kWh; an order-of-magnitude estimate, not a measurement). Footprint broadly rises with capability (Table~\ref{tab:sustainability}) but is driven by \emph{tokens}, not accuracy: under a uniform per-token energy assumption, token-heavy DeepSeek~V3.2 nearly matches Claude's per-case carbon ($6.4$~gCO$_2$) despite its $11.3\times$ lower dollar cost, decoupling the carbon and dollar axes. For an \emph{environmental} benchmark this makes footprint a first-class deployment axis alongside accuracy and cost.

\emph{Build your own:} per-case cost and token counts are logged for every run, so the Pareto frontier and footprint table regenerate from the same committed CSVs when your model set or prices change; the Wh-per-token and grid-intensity constants are single parameters to replace with measured values.

\begin{table}[ht]
  \caption{Estimated energy and carbon per case (tokens $\times\,0.40$~Wh/1k; $400$~gCO$_2$/kWh). Order-of-magnitude estimate; footprint is token-driven.}
  \label{tab:sustainability}
  \centering
  \begin{tabular}{lrr}
    \toprule
    \textbf{Model} & \textbf{Accuracy} & \textbf{gCO$_2$/case} \\
    \midrule
    Claude Sonnet 4 & $60.8\%$ & $6.44$ \\
    DeepSeek V3.2   & $56.3\%$ & $6.42$ \\
    GLM-5           & $50.2\%$ & $5.61$ \\
    Gemini 2.5 Pro  & $48.0\%$ & $4.92$ \\
    GPT-4o          & $41.6\%$ & $3.56$ \\
    Qwen3-235B      & $41.2\%$ & $4.99$ \\
    GPT-OSS-120B    & $34.1\%$ & $4.09$ \\
    Llama 4 Scout   & $26.9\%$ & $3.16$ \\
    Gemma-3-27B     & $15.8\%$ & $3.09$ \\
    \bottomrule
  \end{tabular}
\end{table}

\subsection{Failure Analysis}

\subsubsection{Hard and Model-Dependent Categories}

Several small categories are hard or strongly model-dependent (Figure~\ref{fig:heatmap}; most hold two to four tasks, so these are indicative):

\textbf{Close-value comparison} (\texttt{V5\_21}, the worked example of Section~\ref{sec:benchmark}): most runs never call \texttt{compare\_areas} (Section~\ref{sec:benchmark}); the v6 expansion (Section~\ref{sec:v6}) probes this across a ten-task comparison-heavy set (value gaps $0.6$--$45$~pp). \textbf{Ranking} ($2$ tasks; mean $26\%$, $0$--$67\%$ across models): GLM-5 and GPT-4o ($67\%$) and Claude ($50\%$) handle Top-/Bottom-N province queries while Gemini, GPT-OSS-120B, Llama~4~Scout, and Gemma-3-27B floor at $0\%$, with errors split between wrong tool choice (\texttt{analyze\_area} instead of \texttt{find\_top\_n}) and confabulated rankings. \textbf{Single analysis} ($2$ tasks; $17$--$100\%$) and \textbf{province aggregation} ($2$ tasks; $0$--$75\%$) show similarly wide inter-model variance (Figure~\ref{fig:heatmap}).

Conversely, the easiest categories across the nine models are \textbf{memory/multi-turn} ($6$ tasks, mean $75\%$), \textbf{threshold} ($3$, $72\%$), and \textbf{municipality} ($4$, $69\%$), with \textbf{habitat analysis} ($7$, $65\%$) and \textbf{language} ($6$, $63\%$) close behind: models broadly sustain multi-turn context and parse Galician/Basque inputs but struggle with the multi-step tool composition that ranking and comparison demand (two tasks each, so indicative).

\subsubsection{Failure-mode distribution}

Across all $1{,}429$ failed model--seed--case runs, failures concentrate in tool selection rather than data handling (Table~\ref{tab:failuremodes}). That over two-thirds of all failures are tool-missing---not wrong-data---reframes the difficulty: models fail predominantly by not invoking the right tool, not by mis-reading correct tool output, consistent with uniformly low tool-selection scores (Section~\ref{sec:category}). At the case level, eleven cases are unsolved by \emph{every} model in \emph{every} seed (\texttt{V5\_21}, \texttt{V5\_23}, \texttt{V5\_25}, \texttt{V5\_61}, \texttt{V5\_70}, \texttt{V5\_71}, \texttt{V5\_76}, \texttt{V5\_77}, \texttt{V5\_78}, \texttt{V5\_79}, \texttt{V5\_92}), spanning comparison, deep-dive, tool-selection, and interpretation, though \texttt{V5\_21}'s close-value difficulty is re-attributed to tool composition rather than numeric discrimination in Section~\ref{sec:v6}.

\emph{Build your own:} this taxonomy is a property of the scorer, not of our task set---any case JSON scored by \path{geoagentbench/scoring.py} yields the same first-failing-check distribution, giving you a failure taxonomy for your own domain's tasks for free.

\begin{table}[ht]
  \centering
  \caption{Failure-mode distribution over all $1{,}429$ failed runs (nine models).}
  \label{tab:failuremodes}
  \small
  \begin{tabular}{@{}lr@{}}
    \toprule
    \textbf{Failure mode} & \textbf{\% of failures} \\
    \midrule
    Tool missing (expected tool not called) & $68.0$ \\
    Keyword missing (answer lacks required term) & $19.3$ \\
    Wrong data (incorrect value) & $9.7$ \\
    Rounds exceeded & $1.3$ \\
    Forbidden keyword & $1.3$ \\
    Chart missing & $0.3$ \\
    \bottomrule
  \end{tabular}
\end{table}

\subsection{v6 Expansion: Close-Value Comparison Revisited}
\label{sec:v6}

The close-value finding above rests on a single case (\texttt{V5\_21}), so we ran a ten-task expansion set (\emph{v6}) of comparison-heavy queries over the Portuguese BigEarthNet land-cover layer, spanning ground-truth value gaps from $0.6$ to $45.1$ percentage points. It includes a deliberate large-gap control (\texttt{V6\_05}, $45.1$~pp), two legend-reading tasks, and three aggregation tasks (a threshold count, a sum, and a sum-then-compare). All nine leaderboard models were run under the identical three-seed, temperature-$1.0$ protocol (Table~\ref{tab:v6}).

\begin{figure*}[ht]
    \centering
    \includegraphics[width=0.72\textwidth]{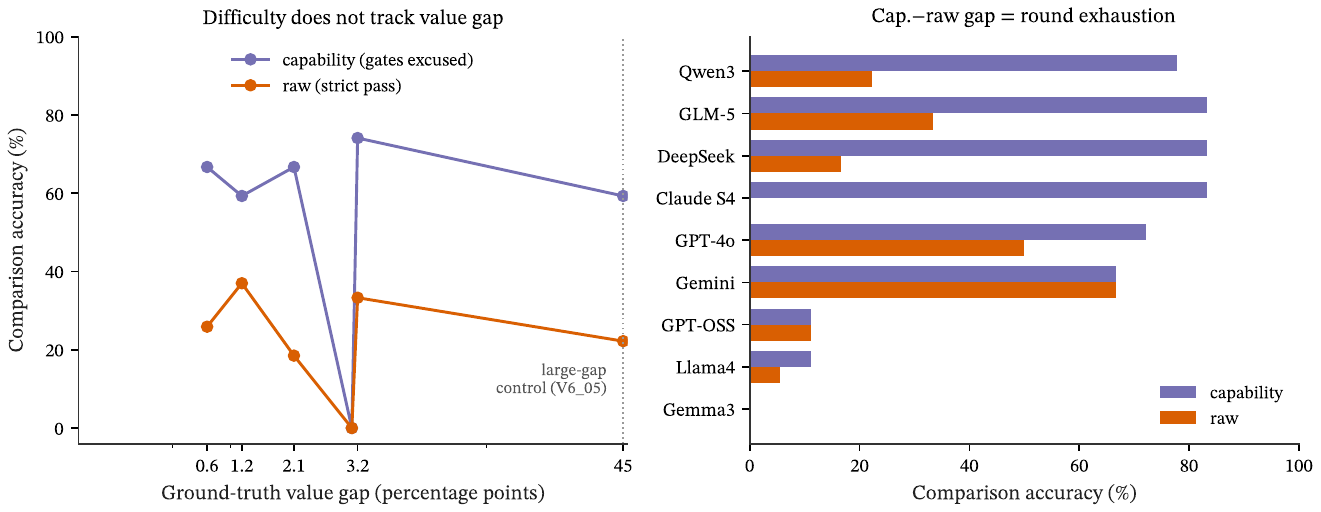}
    \caption{v6 comparison tasks. \emph{Left:} pooled comparison accuracy vs.\ value gap, raw and capability scoring---flat in the gap, and the $45$~pp control is no easier, so difficulty is not near-equal discrimination. \emph{Right:} per-model comparison accuracy, capability vs.\ raw; the bar gap is round/budget exhaustion while composing repeated \texttt{analyze\_area} calls.}
    \Description{}
    \label{fig:v6comp}
\end{figure*}

Three findings refine the v5 result. First, \textbf{the ``unsolved by every model'' statement is not universal}: under strict (raw) scoring Gemini~2.5~Pro solves the majority of comparison runs ($67\%$, including the control at $100\%$) and GPT-4o reaches $50\%$, so at least two models can discriminate near-equal values. Second, \textbf{difficulty does not track the value gap} (Figure~\ref{fig:v6comp}, left): pooled raw accuracy is essentially flat across gaps ($26\%$ at $0.6$~pp, $37\%$ at $1.2$~pp, $19\%$ at $2.1$~pp; one task per gap, so indicative), the large-gap control is \emph{no easier} ($22\%$ raw), and the sum-then-compare task (\texttt{V6\_10}) is the hardest of all ($0\%$ raw across every model and seed). The deficit is therefore not near-equal numeric discrimination but the multi-step tool composition the Portugal layer demands (served only by \texttt{analyze\_area}, each comparison requires several sequential calls). Third, \textbf{this manifests as a raw-versus-capability gap} (Figure~\ref{fig:v6comp}, right): capability scoring---which excuses budget/round exhaustion, whereas the v5 leaderboard excuses cost only---lifts comparison accuracy to $67$--$83\%$ for the top models, so most comparison ``failures'' are timeouts mid-composition rather than wrong answers; Claude~Sonnet~4 is the extreme case at $83\%$ comparison capability but $0/18$ raw. The expansion thus treats the v5 close-value case as indicative rather than a fundamental limitation, re-attributing the residual difficulty to tool composition and turn budget (Figure~\ref{fig:v6comp}).

\paragraph{Combined 103-task metric.} Folding the expansion into the main suite yields the inclusive leaderboard in the last column of Table~\ref{tab:v6}: Claude~Sonnet~4 $61.7\%\pm0.7\%$, DeepSeek~V3.2 $57.9\%\pm3.9\%$, GLM-5 $52.4\%\pm2.6\%$, down to Gemma-3-27B $14.6\%\pm1.0\%$. The v5 ordering is preserved throughout (only Qwen3-235B and GPT-4o swap, within noise), so including the new tasks confirms rather than alters the headline findings; per-suite v6 accuracies remain reported above because the expansion is deliberately comparison-heavy, not distribution-matched. \emph{Build your own:} v6 is the template for hypothesis-driven expansion---a ten-case JSON dropped into \path{geoagentbench/cases/} and registered in one line of \path{case_loader.py} reuses the entire run/score/aggregate pipeline (\path{scripts/aggregate_combined.py} recomputes the combined leaderboard) to probe one targeted hypothesis; any question about your agents can be probed the same way.

\begin{table}[ht]
  \caption{v6 expansion (10 comparison-heavy tasks, 3 seeds) and the combined 103-task suite. \emph{Cap.}\ is capability scoring (budget/round gates excused; the v5 leaderboard excuses cost only); \emph{raw} is strict pass. \emph{103-task comb.}\ is capability accuracy over the full v5+v6 suite (per-seed mean), the inclusive headline metric for this paper.}
  \label{tab:v6}
  \centering
  \small
  \begin{tabular}{lrrrr}
    \toprule
    \textbf{Model} & \textbf{Cap.} & \textbf{Raw} & \textbf{Comp.\ cap/raw} & \textbf{103-task comb.} \\
    \midrule
    Qwen3-235B & $73\%$ & $37\%$ & $78/22\%$ & $44.3\%$ \\
    GLM-5 & $73\%$ & $37\%$ & $83/33\%$ & $52.4\%$ \\
    DeepSeek V3.2 & $73\%$ & $27\%$ & $83/17\%$ & $57.9\%$ \\
    Claude Sonnet 4 & $73\%$ & $20\%$ & $83/0\%$ & $61.7\%$ \\
    GPT-4o & $63\%$ & $50\%$ & $72/50\%$ & $43.7\%$ \\
    Gemini 2.5 Pro & $60\%$ & $60\%$ & $67/67\%$ & $49.2\%$ \\
    GPT-OSS-120B & $20\%$ & $20\%$ & $11/11\%$ & $32.7\%$ \\
    Llama 4 Scout & $17\%$ & $13\%$ & $11/6\%$ & $25.9\%$ \\
    Gemma-3-27B & $3\%$ & $3\%$ & $0/0\%$ & $14.6\%$ \\
    \bottomrule
  \end{tabular}
\end{table}

\subsection{Category-Level Analysis}
\label{sec:category}

\begin{figure}[ht]
    \centering
    \includegraphics[trim={0.1cm 0.2cm 0.1cm 0.6cm}, clip, width=0.88\columnwidth]{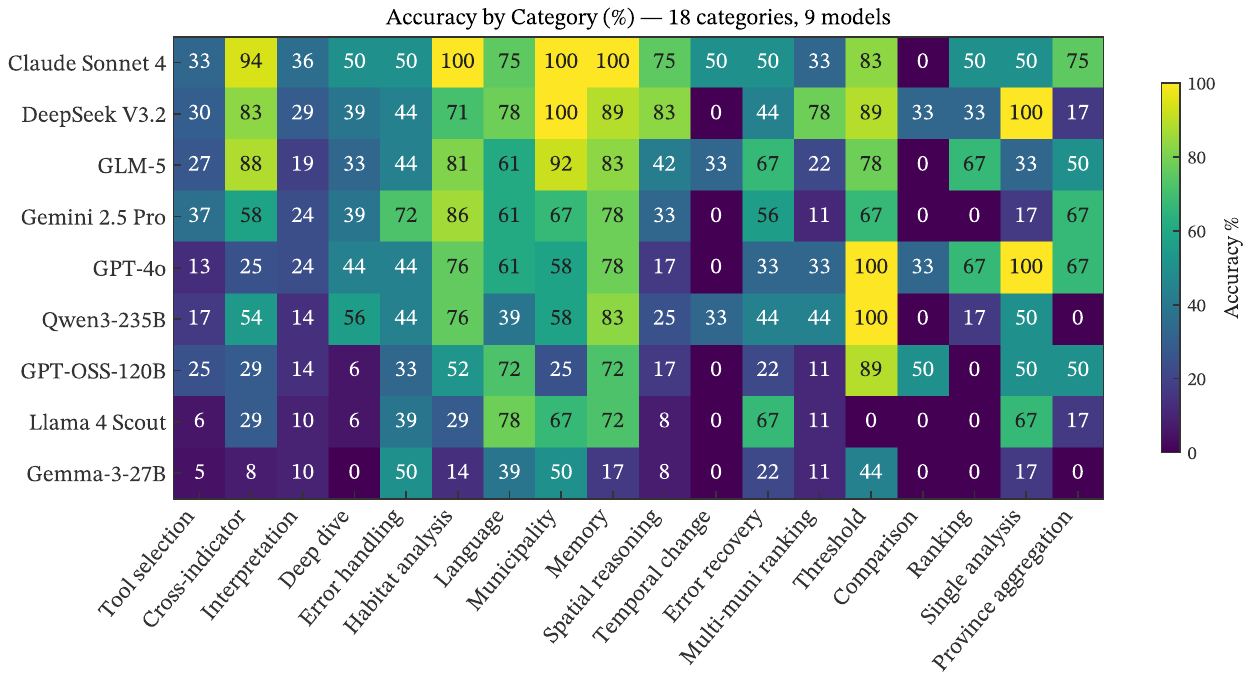}
    \caption{Accuracy by category across the nine models.}
    \Description{}
    \label{fig:heatmap}
\end{figure}

Figure~\ref{fig:heatmap} shows mean accuracy by category across all nine models. Two structural findings emerge:

\textbf{(1) Tool selection, the largest category, is uniformly low.} No model exceeds $37\%$ on its 21 cases: Gemini~2.5~Pro ($37\%$), Claude ($33\%$), DeepSeek ($30\%$), GLM-5 ($27\%$), and GPT-OSS-120B ($25\%$) cluster within a 12-point band; the other four (Qwen3-235B $17\%$ down to Gemma-3-27B $5\%$) trail. \textbf{(2) Model strengths are domain-specialized rather than uniform.} No model dominates every category: Claude leads habitat analysis, memory (both $100\%$), and cross-indicator synthesis ($94\%$); DeepSeek leads spatial reasoning ($83\%$); and GPT-4o co-tops single analysis, threshold (both $100\%$), and ranking ($67\%$) with DeepSeek, Qwen3-235B, and GLM-5 respectively. Each of these leads rests on only $2$--$8$ tasks, so they are indicative rather than statistically separable. A practitioner should consult the per-category breakdown rather than headline accuracy alone. \emph{Build your own:} categories are just a field on each case JSON---define categories that mirror your workload's task patterns and the heatmap and per-category CSVs regenerate for your suite.

\subsection{Prompt-version sensitivity and reproducibility}
\label{sec:prompt}
\label{sec:refresh}

Two controls govern how any tool-calling leaderboard should be read: how much the system prompt moves scores, and whether the harness reproduces. Re-running the v5 suite under the paper's v3 prompt on three current models (single seed) probes both. Moving from the runner's default (v2) prompt to v3 raises Claude~Haiku~4.5 by $+22.6$~pp ($25.8\%\!\to\!48.4\%$) and Gemini~2.5~Pro by $+11.8$~pp ($30.1\%\!\to\!41.9\%$)---a swing larger than every adjacent-rank gap in Table~\ref{tab:leaderboard}, which is why the v3 prompt is held fixed across all models. The fresh Gemini~2.5~Pro run ($41.9\%$) falls inside the paper's three-seed $95\%$ bootstrap interval ($[39.8, 56.3]$, mean $48.0\%$), confirming the harness reproduces to within sampling noise; Claude~Sonnet~4.5 scores $30.1\%$ despite an $85\%$ partial-credit score---a tool-call format mismatch under our scorer (flagged for verification), not reduced capability. These rows are reported alongside, not as a re-ranking of, the accepted nine-model leaderboard. \emph{Build your own:} prompts are versioned files under \path{api/agent/prompts/}---testing your own against the fixed task suite is a new file and a config line, not a code change.

\section{Cross-Dataset Extensibility: BigEarthNet V2}
\label{sec:validation}

To demonstrate GNA's extensibility to a new data domain and country, we incorporate BigEarthNet~V2~\citep{clasen2024reben} as a third data domain covering a different country: the 75k+ Portuguese Sentinel-2 patches are aggregated across 9 mainland districts into a 7-class land cover distribution queryable via the same \texttt{analyze\_area} tool used for Spanish indicators, with fixed numeric ground truth from the pre-computed district statistics enabling deterministic scoring. The 20 BigEarthNet cases span seven task patterns (single-district breakdown, cross-indicator synthesis with Spanish layers, multi-turn recall, chart generation, legend/layer discovery, deep-dive profiling, and two-district comparison); 4 additional cases in other categories bring the total to 24. Aggregated across seeds, \textbf{Claude~Sonnet~4} achieves the highest accuracy on this subset ($100\%$ on the seven dedicated habitat-analysis cases, $94\%$ on the eight cross-indicator cases combining Portuguese land cover with Spanish indicators), followed on the combined subset by \textbf{GLM-5} (habitat $81\%$, cross-indicator $88\%$) and \textbf{DeepSeek~V3.2} (habitat $71\%$, cross-indicator $83\%$), while Gemini~2.5~Pro is second on the habitat cases alone ($86\%$); at seven and eight tasks these orderings are indicative. Beyond the scores, the integration is the framework's extension recipe demonstrated end-to-end. \emph{Build your own:} adding an indicator and a second country required only a JSON statistics file, a prompt update, and new cases---no scoring-engine or protocol changes---and the same path applies to NDVI, precipitation, or your own region's layers.

\section{Limitations and Future Work}
\label{sec:limitations}

\paragraph{Limitations.}

\textbf{Small samples}: with $93$ main-suite tasks across $18$ categories, roughly half hold three or fewer instances, so category- and failure-mode claims are indicative; the aggregate leaderboard is better powered than any single category. \textbf{Serving heterogeneity}: provider-managed stacks with uncontrolled quantization and region (Table~\ref{tab:models}) confound absolute latency and cost, though not the shared tool interface. \textbf{Single-region data}: indicators cover only Spain and Portugal, so geographic generalization is untested. \textbf{Evaluation strictness}: binary pass/fail understates per-step capability (partial credit $71$--$90\%$) and accounts for much of the gap versus looser-scored benchmarks (Section~\ref{sec:positioning}); no prior suite has yet been re-run under our protocol. \textbf{Architecture scope}: every system is a single-agent ReAct loop; multi-agent decompositions deliver large gains on adjacent benchmarks~\citep{luo2025geojsonagents, lee2025geollmsquad}.

\paragraph{Future work.}

Extensions follow directly. \textbf{Scoring}: tolerance-/range-based scoring so near-equal comparisons survive strict binary matching, plus a $\mathrm{pass}^{k}$ reliability metric for production consistency, and re-running prior suites (e.g., GeoBenchX) under GNA's protocol for a same-protocol comparison. \textbf{Coverage}: expanding the model roster, thin categories, task types, regions, and indicators would raise power and test generalization. \textbf{Substrate}: exposing the other MCP primitives (resources, prompts) and subagent-orchestrated clients would broaden the constructs under test. \textbf{Sustainability}: refining the energy/carbon estimate toward measured per-token footprints would make footprint a first-class metric.

\section{Conclusion}
\label{sec:conclusion}

We introduce GeoNatureAgent (GNA), a framework for pre-production evaluation of tool-using agents whose flagship instance benchmarks environmental geospatial analysis through structured tool calling against an open, self-hostable API. Evaluating nine LLMs on $103$ tasks (93-task main suite + ten-task expansion) under three temperature-1.0 seeds:

\begin{enumerate}
    \item \textbf{End-to-end tool orchestration on our environmental suite remains open.} The capability leader (Claude~Sonnet~4) achieves $60.8\% \pm 0.8\%$ on the main suite under strict all-checks scoring (capability-scored $61.7\% \pm 0.7\%$ on all $103$ tasks); only two models exceed $55\%$ and none reaches $65\%$. Per-check partial credit is far higher ($86$--$90\%$ for the top four), so much of the gap to the $85$--$97\%$ reported on general GIS benchmarks reflects our stricter, completion-based scoring (indicative; not a same-protocol comparison).
    \item \textbf{Open-weight models occupy the cost-accuracy Pareto frontier.} Three of the four frontier models are open-weight (Llama~4~Scout, Qwen3-235B, DeepSeek~V3.2).
    \item \textbf{The capability-cost trade-off is steep.} DeepSeek~V3.2 delivers $93\%$ of Claude's capability ($56.3\%$ vs.\ $60.8\%$) at $11.3\times$ lower per-case list-price cost.
    \item \textbf{Systematic failures persist.} Missing tool calls dominate failures ($68\%$), and parameter scale tracks capability only within our small open-weight sample.
\end{enumerate}

GNA is offered as a recipe as much as a result: swap the tool executors behind the MCP server for your domain's operations, write task JSON, and the committed pipeline produces the leaderboard, Pareto frontier, failure taxonomy, and category breakdown for \emph{your} domain. GNA---the MCP tool server, evaluation harness, benchmark, and self-hostable API---is publicly available at \url{https://github.com/gabrielireland/GeoNatureAgent_Benchmark}; the dataset and software are archived at Zenodo~\citep{geonatureagent_dataset, geonatureagent_software}.

\begin{acks}
We thank Darwin Geospatial for infrastructure support and for open-sourcing the code and experiments. The Portuguese land-cover statistics derive from BigEarthNet~V2, which contains modified Copernicus Sentinel data (2017--2018); we acknowledge the European Space Agency and the Copernicus programme.
\end{acks}

\bibliographystyle{ACM-Reference-Format}
\bibliography{references}

\end{document}